\RequirePackage{fix-cm}
\documentclass[smallcondensed]{svjour3}     
\smartqed  
\usepackage{graphicx}
\usepackage{subcaption}
\usepackage{amsmath}
\usepackage{algpseudocode}
\usepackage{algorithm}
\algnewcommand\algorithmicforeach{\textbf{foreach}}
\algdef{S}[FOR]{ForEach}[1]{\algorithmicforeach\ #1\ \algorithmicdo}
\let\oldReturn\Return
\renewcommand{\Return}{\State\oldReturn}
\usepackage{caption}
\usepackage{enumitem}
\usepackage{url}
\usepackage{float}
\newcommand{\subparagraph}{}

\begin{document}

\title{AnyThreat: An Opportunistic Knowledge Discovery Approach to Insider Threat Detection}


\titlerunning{An Opportunistic Knowledge Discovery Approach to Insider Threat Detection}        

\author{Diana Haidar         \and
        Mohamed Medhat Gaber \and
        Yevgeniya Kovalchuk 
}


\institute{D. Haidar \at
              Birmingham, United Kingdom \\
              \email{diana.haidar@mail.bcu.ac.uk}           
           \and
           M. M. Gaber \at
              Birmingham, United Kingdom
           \and
          Y. Kovalchuk \at
              Birmingham, United Kingdom
}


\maketitle

\begin{abstract}
Insider threat detection is getting an increased concern from academia, industry, and governments due to the growing number of malicious insider incidents. The existing approaches proposed for detecting insider threats still have a common shortcoming, which is the high number of false alarms (false positives). The challenge in these approaches is that it is essential to detect all anomalous behaviours which belong to a particular threat. To address this shortcoming, we propose an opportunistic knowledge discovery system, namely AnyThreat, with the aim to detect any anomalous behaviour in all malicious insider threats. We design the AnyThreat system with four components. (1) A feature engineering component, which constructs community data sets from the activity logs of a group of users having the same role. (2) An oversampling component, where we propose a novel oversampling technique named Artificial Minority Oversampling and Trapper REmoval (AMOTRE). AMOTRE first removes the minority (anomalous) instances that have a high resemblance with normal (majority) instances to reduce the number of false alarms, then it synthetically oversamples the minority class by shielding the border of the majority class. (3) A class decomposition component, which is introduced to cluster the instances of the majority class into subclasses to weaken the effect of the majority class without information loss. (4) A classification component, which applies a classification method on the subclasses to achieve a better separation between the majority class(es) and the minority class(es). AnyThreat is evaluated on synthetic data sets generated by Carnegie Mellon University. It detects approximately $87.5$\% of malicious insider threats, and achieves the minimum of false positives$\text{=}3.36$\%.

\keywords{Knowledge Discovery \and Feature Engineering \and Oversampling \and Class Decomposition \and Classification \and Insider Threat Detection}
\end{abstract}

\section{Introduction} 
\label{sec:1} 
Insider threat detection is a growing challenge for companies and governments trying to protect their assets from malicious insider threats. An insider is a current or former employee, contractor, or business partner of an organisation who has authorised access to the network, system, or data \cite{cappelli2012cert}. A malicious insider threat refers to a malicious insider who exploits their privileges with the intention to compromise the confidentiality, integrity, or availability of the system or data \cite{nurse2014critical}. 

The 2018 Insider Threat Report \cite{crowd2018report}, released by the Crowd Research Partners, found that 90\% of the organisations feel vulnerable to insider attacks, where 53\% of the survey respondents confirmed typically less than five insider attacks against their organisation in the past year. Furthermore, the 2018 Insider Threat Report states that 64\% of the organisations are focusing on the detection of the insider threats followed by the prevention and post breach forensics.

The insider threat problem represents a unique knowledge discovery problem that has been addressed either as a sequence-based problem, or as a behaviour-based problem. The sequence-based approaches define an insider threat as a set of behaviours (i.e. actions, commands) executed in a specific order of time by a malicious insider. The behaviour-based approaches define an insider threat as a set of behaviours executed by a malicious insider, regardless of the time order. The existing approaches still have a common shortcoming when addressing the insider threat problem, which is the high number false alarms. The challenge in these approaches is that it is essential to detect \textbf{all} behaviours which belong to a particular threat. A malicious insider threat is devised of a complex pattern of anomalous behaviours carried out by a malicious insider, which makes it difficult to detect all behaviours attributed to all malicious insider threats. 

To overcome the shortcoming of the high number of false alarms and to address the challenge in the existing approaches, we propose a knowledge discovery system, namely AnyThreat, with the aim to detect any-behaviour-all-threats; it is sufficient to detect \textbf{any} anomalous behaviour in \textbf{all} malicious insider threats. In other words, we can \textbf{hunt} a malicious insider threat by at least detecting one anomalous behaviour among the anomalous behaviours associated to this threat. Designing the knowledge discovery system with such a relaxing condition will contribute in reducing the false alarms. We call this approach \textit{opportunistic}, or sometimes we refer to it as \textit{threat hunting}. 

We design the knowledge discovery system AnyThreat with four components: a feature engineering component, an oversampling component, a class decomposition component, and a classification component. We elaborate on the contribution of each component in the following.

\begin{itemize}[noitemsep]
\item The \textbf{feature engineering component} preprocesses the system and network logs of users behaviour and extracts the feature set to define the feature space of the insider threat problem. 

\item With the scarcity of instances for `anomalous' behaviours (attributed to malicious insider threats) in an organisation and the abundance of instances for `normal' behaviours, we shape the insider threat problem as an \textit{opportunistic classification problem} with \textit{class imbalance}. The \textbf{oversampling component} tackles the class imbalance in the insider threat data, where the minority class consists of the rare anomalous behaviours that map to different scenarios of threats, and the majority class consists of the normal behaviours of a community of users (i.e. a group of users having the same role). We devised an oversampling technique, namely Artificial Minority Oversampling and Trapper REmoval (AMOTRE), for the oversampling component. AMOTRE implements a Trapper REmoval (-TRE) method to remove behaviours from the insider threat data (minority class) that have high resemblance with the normal behaviour, and can be a trapper for the classifier to generate false alarms. The removal of such instances can be mostly safe, as based on the opportunistic approach, the detection of all behaviours carried out by the insider is not required. AMOTRE then synthetically oversamples the minority class by shielding the border of the majority class. 
It is worth noting that the proposed AMOTRE technique can be replaced by any other oversampling technique such as Synthetic Minority Oversampling Technique (SMOTE) \cite{chawla2002smote} and its variations, which are the most successful in this area.

\item Class decomposition is the process of using clustering over the instances of a class or more in a data set to detect subclasses \cite{vilalta2003class}. The process can be applied to all classes or a subset of the classes present in the data. In this system, the \textbf{Class Decomposition (CD) component} is used to weaken the effect of the majority class \textit{without information loss} or \textit{cluster-based sampling}.

\item The \textbf{classification component} applies a classification method to delineate one (or multiple) decision boundaries to separate the majority instances and the minority instances, thus improving the prediction of new instances.

\end{itemize}

The rest of the article is organised as follows. In Section \ref{sec:2}, we give a review of the sequence-based approaches and the behaviour-based approaches proposed for insider threat detection. In Section \ref{sec:3}, we present the knowledge discovery system AnyThreat, and we describe and formulate the four components. In Section \ref{sec:4}, we evaluate the performance of the proposed opportunistic AnyThreat system utilising the proposed the state-of-the-art SMOTE or the proposed AMOTRE oversampling technique with and without the class decomposition variations. Finally, we conclude this article with a conclusion and future work in Section \ref{sec:5}.

\vspace{-1em}
\section{Related Work} 
\label{sec:2} 
There exists a significant body of research on applying data mining approaches for detecting insider threats. In this Section, we discuss related work on the sequence-based approaches and the behaviour-based classification approaches proposed for insider threat detection.

\vspace{-1em}

\subsection{Sequence-based Approaches}
\label{sec:2,subsec:1} 
The sequence-based approaches define a malicious insider threat as a sequence of behaviours executed in a specific order attributed to a malicious insider. In the following, we provide a review of the proposed sequence-based approaches for detecting insider threats.

A recent framework based on a graph approach and isolation Forest (iForest) was presented by Gamachchi et al. \cite{gamachchi2017graph} to isolate suspicious malicious insiders from the workforce. Furthermore, Gamachchi and Boztas \cite{gamachchi2017insider} propose a framework based on attributed graph clustering and outlier ranking for insider threat detection. The proposed framework shares some characteristics with the author's framework \cite{gamachchi2017graph}, however, an outlier ranking technique named GOutRank is employed instead of iForest to detect the malicious users. Moriano et al. \cite{moriano2017insider} suggest a temporal bipartite graph of user-system interactions to detect anomalous time intervals. Eberle and Holder \cite{eberle2011insider} suggest a graph-based framework to detect malicious insider threats using graph substructures. The framework employs three different versions of the Graph-Based Anomaly Detection (GBAD) algorithm: information theoretic GBAD, probabilistic GBAD, and maximum partial substructure GBAD. Huang and Stamp \cite{huang2011masquerade} propose Profile Hidden Markov Model (PHMM) compared to the standard Hidden Markov Model (HMM) to detect masqueraders (i.e. insiders who infiltrate a system’s access control to overcome access to a user’s account). In contrast to HMM, PHMM relies on positional observations which include a session start and a session end features. Tang et al. \cite{tang2009insider} suggest a hybrid approach of Dynamic Bayesian Network (DBN) and HMM. The hybrid approach models data over session slots and assigns an abnormality score for each user.

\subsection{Behaviour-based Classification Approaches}
\label{sec:2,subsec:2} 
The behaviour-based approaches defines a malicious insider threat as a set of behaviours (instances). An instance is defined as a feature vector (a set of features) extracted from the activity logs of a user or a group of users. Hence, an anomalous behaviour is a feature vector whose features' values deviate from the normal baseline. In what follows, a review of the behaviour-based classification approaches proposed for insider threat detection is provided.

Mayhew et al. \cite{mayhew2015use} describe a Behaviour-Based Access Control (BBAC) approach that analyses user behaviour at the network layer, the HTTP request layer, and the document layer to detect behaviour changes. BBAC combines Support Vector Machine (SVM), $k$-means clustering, and C4.5 decision tree to improve scalability and reduce false positives. Punithavathani et al. \cite{punithavathani2015surveillance} present an Insider Attack Detection System (IADS) based on the supervised $k$-Nearest Neighbours ($k$-NN) to counter insider threats in critical networks. Azaria et al. \cite{azaria2014behavioral} present a Behavioural Analysis of Insider Threat (BAIT) framework that contains bootstrapping algorithms built on top of SVM or Naives Bayes classifiers. Gates et al. \cite{gates2014detecting} use the structure of the file system hierarchy to measure the level of access similarity of the files, and detect anomalous behaviour based on a predefined threshold. The paper defines access similarity measure techniques including, self score which compares a user's access similarity to a user's historical accesses; and a relative score which compares a user's average score to other users' accesses. Axelrad et al. \cite{axelrad2013bayesian} present a Bayesian Network (BN) approach to predict insider threats based on psychological features of malicious insiders. Barrios \cite{barrios2013multi} develops a multi-level framework, called database Intrusion Detection System (dIDS), that employs BN to detect malicious transactions.

\vspace{-1em}

\subsection{The Shortcoming of High Number of False Alarms}
\label{sec:2,subsec:3} 
The reviewed approaches have shown merit in addressing the insider threat detection problem, however, they still have a common shortcoming which is the high number of false alarms. Some of these approaches report a low FP rate, such as: FP rate$\text{=}0.18-1\%$ in BBAC \cite{mayhew2015use}; FP rate$\text{=}2.5\%$ in \cite{gates2014detecting}; and FP rate$\text{=}4.73\%$ in BAIT \cite{azaria2014behavioral}. However, these approaches are evaluated on data sets not specifically designed for insider threats. Thus, their use does not constitute the challenges of variety and complexity in threat scenarios. 

To address this shortcoming, in this work, we propose an opportunistic knowledge discovery system, namely AnyThreat, to tackle the insider threat problem. To our knowledge, none of the existing approaches addressed the insider threat problem from the perspective of class imbalance, which was discussed in \cite{azaria2014behavioral}. AnyThreat implements the first class imbalance data approach for insider threat detection. We investigate how the concept of class decomposition and the oversampling of a selective set of anomalous instances address the class imbalance data problem and minimise the number of false alarms, as detailed in the following Section. 

\vspace{-1em}
\section{Proposed Opportunistic Knowledge Discovery System}  
\label{sec:3} 

\begin{figure}[!t]
\scriptsize
\captionsetup{font=scriptsize}
	\centering
	\includegraphics[width=\textwidth]{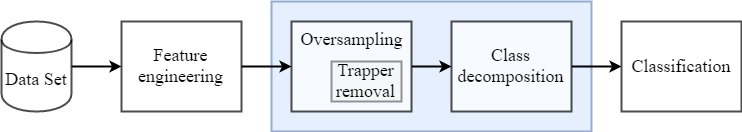}
	\caption{Workflow of the Proposed Opportunistic Knowledge Discovery System -- AnyThreat.}
	\label{fig:anythreatsystem}
\end{figure}

In this section, we present the proposed opportunistic knowledge discovery system, namely AnyThreat, with the aim to detect any-behaviour-all-threat. It is sufficient to detect any anomalous behaviour of all malicious insider threats in the data set. Fig. \ref{fig:anythreatsystem} illustrates the workflow of the AnyThreat system. Given that the data of both `normal' behaviours (instances) and `anomalous' behaviours (attributed to malicious insider threats) are available in an organisation, the AnyThreat system takes as input the community users behaviours from a data set (database). We design the AnyThreat system with four components: a feature engineering component, an oversampling component, a class decomposition component, and a classification component. In the following, we elaborate on the role of each component in the system workflow. 

\subsection{Feature Engineering Component}
\label{sec:3,subsec:1}
The first step to tackle the insider threat problem in the AnyThreat system is to identify the insider threat feature space. The role of the feature engineering component is to preprocess the data set, and define and extract the feature set to prepare the data for the data mining approach. In this work, we utilised the $r5.2$ release of the synthetic data sets generated by Carnegie Mellon University - Community Emergency Response Team (CMU-CERT) \cite{glasser2013bridging,certinsider}, which implements a variety of malicious insider threat scenarios. The $r5.2$ CMU-CERT data sets consist of system and network logs for the activities carried out by users in an organisation over the course of $18$ months (e.g. logons, connecting removable devices, copying files, browsing websites, sending emails, etc.). Based on the literature \cite{bose2017detecting,legg2017automated}, we extract a feature set from these logs to represent the baseline of users' behaviour. We categorise the features into five groups as follows:

\begin{itemize}[noitemsep]
\item Frequency-based `$integer$': assess the frequency of an activity carried out by the users in a specified community during a defined period of time (e.g. frequency of logon, frequency of connecting devices);
\item Time-based `$integer$': assess an activity carried out within the non-working hours (e.g. logon after work hours, device usage after work hours);
\item Boolean `$flag \text{=} \lbrace 0,1 \rbrace$': assess the presence/absence of an activity-related information (e.g. non-empty email-bcc, a non-employee email recipient, sensitive file extension);
\item Attribute-based `$integer$': are more specialised features which assess an activity with respect to a particular value of an attribute (e.g. browsing a particular URL \textit{job websites, WikiLeaks}); and
\item Others `$integer$': assess the count of other activity-related information (e.g. number of email recipients, number of attachments to emails).
\end{itemize}

Based on the identified feature set, we construct community data sets, where a community data set represents the behaviour of a group of users having the same role (e.g. Salesman, IT admin) over session slots. A session slot defines the period of time from \textit{start time} to \textit{end time}, such that the behaviour logs of all users in the community during this period of time are used to extract a vector of feature values (i.e. instance). In this work, the session slot is defined per $4$ hours, which is long enough to extract an instance that provides an adequate evidence of anomalous behaviour. If the session slot is chosen per minutes, for example, the extracted instances would lack adequate evidence of the occurrence of anomalous behaviour. On the other hand, if the session slot is chosen per days/weeks, for example, the period of time will be too long to capture the anomalous behaviour blurred among the normal behaviour in the extracted vector of feature values.

Among the $2000$ employees in the $r5.2$ data sets, we extracted the data logs for the users (employees) belonging to the following three community data sets to be later utilised to validate the experiments:
\begin{itemize}[noitemsep]
\item Production line worker (com-P): It consists of $300$ users, including $17$ malicious insiders. It has the scenarios $\lbrace s1,s2,s4 \rbrace$ implemented;
\item Salesman (com-S): It consists of $298$ users, including $22$ malicious insiders. It has the scenarios $\lbrace s1,s2,s4 \rbrace$ implemented; and
\item IT admin (com-I): It consists of $80$ users, including $12$ malicious insiders. It has the scenarios $\lbrace s2,s3 \rbrace$ implemented.
\end{itemize}     

After constructing a community data set, we normalise each vector of feature values to the range $\left[  0, 1 \right] $, and associate it with a class label $\lbrace Normal, Anomalous \rbrace$.

\vspace{-1em}

\subsection{Oversampling Component}
\label{sec:3,subsec:2}
In the oversampling component, we tackle the class imbalance data problem by sampling the instances of the minority class to modify the original distribution of data among the classes and to achieve an approximate balance between the majority class (or clusters of the majority class) and the minority class. 

A leading and widely adopted sampling technique is a hybrid technique, called Synthetic Minority Oversampling Technique (SMOTE) \cite{chawla2002smote}. In the following, we introduce SMOTE as an oversampling technique that can be employed in the oversampling component of the AnyThreat system. We then present, describe, and formalise the proposed AMOTRE oversampling technique.

\vspace{-1em}

\subsubsection{SMOTE Oversampling Technique}
\label{sec:3,subsec:2,subsubsec:1}
The SMOTE technique was first proposed as an oversampling technique, which introduces artificial samples into the minority class, making it more dense. Let $I$ represent the original set of instances that belong to the minority class, and $perc.over$ be the percentage of oversampling. For each instance $A^{t} \in I$, SMOTE finds its $k_{SMOTE}$ (parameter $k_{SMOTE}$) nearest neighbours from the set of minority class instances $I$. Then, based on the parameter $perc.over$, a $perc.over / 100$ number of artificial samples is generated for each $A^{t}$. An artificial sample $S^{t}$ for $A^{t}$ is introduced along the segment line joining $A^{t}$ and any of its $k_{SMOTE}$ nearest neighbours (randomly selected). In this way, for each $A^{t}$, the artificial samples will be generated along the segments joining any/all of its $k_{SMOTE}$ nearest neighbours. Consider, for example, if $perc.over=200$, then SMOTE generates $perc.over / 100 \text{=} 2$ artificial samples for each $A^{t}$. Thus, it introduces $(perc.over / 100) \times card(I) \text{=} 2 \times card(I)$ samples into the minority class.

Undersampling was then integrated into the SMOTE technique to remove random samples from the majority class, so that the minority class becomes a specified percentage of the majority class. Let $perc.under$ represent the percentage of undersampling. For instance, if $perc.under=300$ and the number of artificial samples added to the minority class is $60$, then only $(perc.under /100) \times 60 \text{=} 180$ majority instances are randomly selected to remain in the set of majority class instances. The other majority instances are removed. 
In this way, the SMOTE technique reverses the initial bias of the classifier towards the majority class in the favour of the minority class.

SMOTE \cite{chawla2002smote} and its variations are the most successful in this area. In the following, we describe and argue the suitablilty of some of the techniques proposed as an extension for SMOTE \cite{han2005borderline,bunkhumpornpat2009safe,maciejewski2011local,barua2014mwmote}. For instance, Borderline-SMOTE \cite{han2005borderline} seeks to oversample only the \textit{borderline} minority instances; those which are in the borderline areas (i.e. on or near the decision boundary), where the majority class and the minority class overlap. This technique generates synthetic samples in the neighbourhood of the borderline(s), where the minority instances are most likely to be misclassified. We argue that Borderline-SMOTE is not suitable for the insider threat problem due to the following. First, this technique assumes a well-defined border(s) of the minority class along all the dimensions (i.e. features), which is not applicable in the insider threat problem. The complexity of the malicious insider threat scenarios manifests in the high similarity of the anomalous behaviours to normal behaviours. The minority instances (i.e. anomalous behaviours) are typically similar to the majority instances (i.e. normal behaviours), however, at the level of \textit{some} features. Second, the concept of the Borderline-SMOTE disregards the oversampling of minority instances within the minority class, and therefore the creation of clusters for the minority class is not applicable, making the process of class decomposition (later described in Section \ref{sec:3,subsec:3}) less effective.                                                                                              
Hence, in this work, we consider the traditional SMOTE as a more suitable technique for insider threat problem, and accordingly, we use it as a benchmark for our experimental study, detailed later in Section \ref{sec:4}.
 
\vspace{-1em}

\subsubsection{Proposed AMOTRE Oversampling Technique}
\label{sec:3,subsec:2,subsubsec:2}
In SMOTE oversampling technique, the process of generating new instances is solely dependent on existing instances in the minority class, or defines the border instances to oversample based on nearest neighbours using all dimensions collectively. In the following, we propose a selective oversampling technique, namely AMOTRE, as an alternative to the state-of-the-art SMOTE sampling technique to employ in the oversampling component of the AnyThreat system. In AMOTRE, and for the first time, we constrain the generation of new instances when synthetically oversampling the minority class by shielding the borders of the majority class \textit{along each dimension (i.e. feature) separately} in the data set, preventing the generation of instances that may be positioned in close proximity to the instances of the majority class that in turn increases the false alarms.

In the following, we describe the steps of the AMOTRE oversampling technique, and we later evaluate its performance in Section \ref{sec:4}.
	
\paragraph{Identifying the Peculiarity of Minority Instances} 
The first step in the AMOTRE oversampling technique is to identify the peculiarity of the minority instances. This guides the process of generating artificial samples for each minority instance $A^{t}$ in the minority class $I$. Algorithm \ref{alg:peculiarminority} gives a brief formalisation for identifying the peculiarity of the minority instances. 

\begin{algorithm}[!t]
\caption{Identifying peculiarity of minority instances.}
\label{alg:peculiarminority}
\begin{algorithmic}[1]
    \ForEach{$A^{t} \in I$}
    	\State compute $perclof_{M}^{t}$
    	\State compute $perclof_{I}^{t}$
    	\If{$perclof_{M}^{t} < \tau$}
    		\State remove \textit{trapper} instance $A^{t}$ from $I$
    		\State $I_{r} \gets I \setminus A^{t}$
    	\EndIf
    
    \EndFor
    \ForEach{$A^{t} \in I_{r}$}
    	\State $p^{t} = \left( \lceil perclof_{M}^{t} \rceil \times \lceil perclof_{I}^{t} \rceil \right) / 10^{4} $
    	\State $p \gets p \cup p^{t}$
   	\EndFor
	\Return $p$,$I_{r}$
\end{algorithmic}
\end{algorithm}

\subparagraph{\emph{LOF for Minority Instances}}
We utilise the Local Outlier Factor (LOF) \cite{breunig2000lof} method to find the LOF for each minority instance $A^{t} \in I$. 
LOF is a density-based method that calculates the local outlier factor (score) for each instance in a data set with respect to its $k_{LOF}$ (parameter $k_{LOF}$) nearest neighbours from the whole data set. For instance, consider that LOF is normalised in the range $[0,1]$. If the location of an instance is in a high-density region (cluster), then the LOF is closer to $0$ (too low). However, if the instance is in a low-density region, then the LOF is closer to $1$ (too high). The advantage of LOF compared to global outlier methods is that LOF can identify local outliers in certain regions of the data set which would not be identified as outliers with respect to the whole data set.

We introduce two modified versions of LOF to utilise in the approach proposed in this article: $lof_{M}^{t}$ and $lof_{I}^{t}$. In both versions, the idea is to calculate the local outlier factor for each minority instance $A^{t} \in I$ in the training data set. $lof_{M}^{t}$ and $lof_{I}^{t}$ are tuned for different values of $k_{LOF}$. 

We define $lof_{M}^{t}$ and $lof_{I}^{t}$ as follows:
\begin{definition}{\boldmath $lof_{M}^{t}$}
\normalfont The $lof_{M}^{t}$ is the LOF for a minority instance $A^{t}$ with respect to the $k_{LOF}$ nearest neighbours from the majority class instances ($M$) only excluding the other minority class. In $lof_{M}^{t}$, we select the value of $k_{LOF}=\sqrt{1+card(M)}$ (thumb-rule), where the radicand $1+card(M)$ represents the number of instances utilised to calculate $lof_{M}^{t}$ for $A^{t}$ ($1$ minority instance $A^{t}$ + all of majority instances $M$).
\end{definition}

\begin{definition}{\boldmath $lof_{I}^{t}$}
\normalfont The $lof_{I}^{t}$ is the LOF for a minority instance $A^{t}$ with respect to the $k_{LOF}$ nearest neighbours from the minority class instances ($I \setminus A^{t}$) only excluding the majority class. In $lof_{I}^{t}$, we select the value of $k_{LOF}=\sqrt{card(I)}$ (thumb-rule), where the radicand $card(I)$ represents the number of minority instances utilised to calculate $lof_{I}^{t}$ for $A^{t}$ (all minority instances $I$ including $A^{t}$).  
\end{definition}

Note that $lof_{M}^{t}$ is still calculated with respect to the whole majority class $M$ (i.e. all the majority instances) regardless of whether class $M$ is decomposed into clusters (subclasses). The reason behind this is that class decomposition only improves classification (from \textit{two-class} to \textit{multi-class} classification) to weaken the effect of the majority class. Similarly, $lof_{I}^{t}$ is calculated with respect the the whole minority class $I$ (i.e. all other minority instances) regardless if class $I$ is decomposed into clusters (subclasses).

\begin{figure}[!t]
\scriptsize
\captionsetup{font=scriptsize}
	\centering
	\includegraphics[width=.6\textwidth]{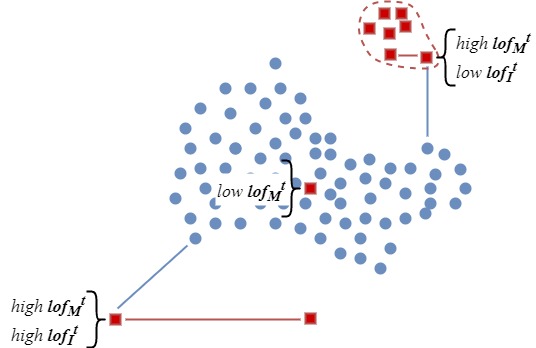}
	\caption{$lof_{M}^{t}$ and $lof_{I}^{t}$ cases of insider threat instances.}	
	\label{fig:LOF}
\end{figure}

In the following, we infer the degree of outlierness of a minority instance $A^{t}$ based on the values of both $lof_{M}^{t}$ and $lof_{I}^{t}$ in three different cases illustrated in Figure \ref{fig:LOF}:
\begin{itemize}[noitemsep]
\item $high \ lof_{M}^{t}$ (close to 1) and $low \ lof_{I}^{t}$ (close to 0) $\Rightarrow A^{t}$ is in a high-density region of minority instances away from the majority instances.
\item $high \ lof_{M}^{t}$ (close to 1) and $high \ lof_{I}^{t}$ (close to 1) $\Rightarrow A^{t}$ is in an outlier away from the majority and minority instances. We call $A^{t}$ here an \textit{extreme outlier} instance.
\item $low \ lof_{M}^{t}$ (close to 0) $\Rightarrow A^{t}$ is located in a high-density region of majority instances.
\end{itemize}

\subparagraph{\emph{The Percentile Rank for Minority Instances}}
The inference from the values of $lof_{M}^{t}$ and $lof_{I}^{t}$ in each of the above mentioned cases controls the number of artificial samples to be generated per minority instance. 
In $lof_{M}^{t}$, we define $perclof_{M}^{t}$ as the percentile rank for each minority instance $A^{t}$ compared to the majority instances. For instance, if the $lof_{M}^{t}$ for $A^{t}$ is the highest compared to all the majority instances (i.e. $lof_{M}^{t} \text{=} 1$), then $perclof_{M}^{t}=100$. If $lof_{M}^{t}\text{=} 0.7$, then $perclof_{M}^{t} \text{=} 70$; the $lof_{M}^{t}$ for $A^{t}$ is greater than $70\%$ of the majority instances. Similarly, we define $perclof_{I}^{t}$ as the percentile rank for each minority instance $A^{t}$ compared to other minority instances.

Removing Trapper Instances:
The anomalous behaviours attributed to malicious insiders often have a high resemblance with the normal behaviours of the insiders or their community (i.e. a group of users having the same role). In the feature space, this appears as minority (anomalous) instances located in a high-density region of majority (normal) instances. These minority instances are characterised by a low degree of outlierness (i.e. $low \ lof_{M}^{t}$ (close to 0)), thus a $low \ perclof_{M}^{t}$. 

We define the parameter $\tau$ to be the survival threshold for the minority instances, such that each minority instance $A^{t} \in I$ having a $perclof_{M}^{t} < \tau$ is considered as a \textit{trapper instance}. The high resemblance of the trapper instances with the surrounding majority instances, would \textit{trap} the classifier to detect the surrounding majority instances as minority (FPs). In our AMOTRE oversampling technique, we propose to take a precautionary step and remove these trapper instances, in an attempt to reduce its contribution to flagging false alarms (FPs). Hence, a new set of minority instances $I_{r}$ will include the minority instances having $perclof_{M}^{t} \geq \tau; \forall A^{t}$ excluding the removed \textit{trapper instances} as detailed in Algorithm \ref{alg:peculiarminority}.

The idea of removing trapper instances is supported by the aim of to detect any-behaviour-all-threats in the opportunistic approach, where it is sufficient to detect any anomalous behaviour, not necessarily all behaviours, for each malicious insider threat. This means that removing a minority trapper instance from the set $I$ is practically removing an anomalous behaviour from all the anomalous behaviours associated to a malicious insider threat. In other words, the \textit{rest} of the anomalous behaviours associated with the malicious insider threat still exist, which permit the detection of the threat (i.e. insider) regardless of the removed anomalous behaviour. Therefore, removing trapper instances not only can improve the performance of the classifier, but also is supported by our ultimate aim to detect any-behaviour-all-threat.

\subparagraph{\emph{Peculiarity of Remaining Minority Instances}}
We define the peculiarity of a minority instance $A^{t} \in I_{r}$ using the probability $p^{t}$. $p^{t}$ represents the probability of generating artificial samples for $A^{t}$. Consider the Eq. \ref{eq:1}:

\begin{equation}
\label{eq:1}
p^{t} = \left( \lceil perclof_{M}^{t} \rceil \times \lceil perclof_{I}^{t} \rceil \right) / 10^{4} 
\end{equation}
where $p^{t}$ is a probability devised from the product of $perclof_{M}^{t}$ and $perclof_{I}^{t}$. The rationale behind using the product is to utilise the degree of outlierness of the instance $A^{t}$ with respect to both: (1) the majority instances ($perclof_{M}^{t}$), and (2) the minority instances ($perclof_{I}^{t}$). In other words, $p^{t}$ actually determines the peculiarity of an instance $A^{t}$ based on its location among the data distribution of both majority instances and minority instances. And accordingly, $p^{t}$ gives the probability of generating artificial samples for $A^{t}$. 

\paragraph{Generating Artificial Samples for Minority Instances} 

Let $perc.over$ represent the percentage of artificial samples to be generated, and let $numS = (perc.over / 100) \times card(I_{r})$ represent the number of artificial samples to be generated. In the following, we only consider the set of remaining minority instances $I_{r}$, excluding the removed trapper instances, to generate the artificial samples. As aforementioned, the peculiarity of the minority instances, which manifests in the probability $p^{t}$ for each $A^{t} \in I_{r}$, will guide the process of sampling. The steps described below are repeated for a number of iterations until $numS$ of artificial samples is generated. Algorithm \ref{alg:artificialsampling} gives a brief formalisation for generating artificial samples for minority instances.

\begin{algorithm}[!t]
\caption{Generating an artificial sample $S^{t}$.}
\label{alg:artificialsampling}
\begin{algorithmic}[1]

   	\State $numS \gets (perc.over / 100) \times card(I_{r})$
	\While{$card(I_{s}) < numS$}
		\State $I_{c} \sim D_{p}$
		\ForEach{$A^{t} \in I_{c}$}
			\State $S^{t} \gets$ \Call{generateSample}{$A^{t}$,$M$}
			\State $I_{s} \gets I_{s} \cup S^{t}$
		\EndFor
	\EndWhile
	\Return $I_{s}$
	\\

\Function{generateSample}{$A^{t}$,$M$}
    
    \ForEach{$a^{t}_{f} \  in \ A^{t}$}
		\State $pos \ n^{t'}_{f} \gets$ positive nearest neighbour in $M$ at the level of feature $f$	
		\State $neg \ n^{t'}_{f} \gets$ negative nearest neighbour in $M$ at the level of feature $f$
		\State $dirS \sim D_{prob^{+}}$
		\If{$\exists pos \ n^{t'}_{f} \land neg \ n^{t'}_{f}$}
			\State $dirN \gets dirS$\;
			\If{$dirS = +1$}
				\State $distN \gets dist(a^{t}_{f}, pos \ n^{t'}_{f})$
			\Else
				\State $distN \gets dist(a^{t}_{f}, neg \ n^{t'}_{f})$
			\EndIf
					
		\EndIf
		\If{$\exists pos \ n^{t'}_{f}$}
			\State $distN \gets dist(a^{t}_{f}, pos \ n^{t'}_{f})$
		\EndIf
		\If{$\exists neg \ n^{t'}_{f}$}
			\State $distN \gets dist(a^{t}_{f}, neg \ n^{t'}_{f})$
		\EndIf

		\State $s^{t}_{f} \gets a^{t}_{f} \ + dirS \times rand(0:\lambda \times distN)$
		\State $S^{t} \gets S^{t} \cup s^{t}_{f}$		
	\EndFor
	\Return $S^{t}$
\EndFunction
\end{algorithmic}
\end{algorithm}

\subparagraph{\emph{Identifying the Chances for Sampling Minority Instances Per Iteration}}
A probability distribution $D_{p}$ is devised using the above probabilities $p^{t} \ \forall A^{t} \in I_{r}$ to determine whether an artificial sample $S^{t}$ will be generated for $A^{t}$ at the current iteration. We define $I_{c} \sim D_{p}$ where $I_{c}$ represents the set of instances chosen according to $D_{p}$ to be sampled, and $p$ is a continuous range of the values of $p^{t}$. For instance, if $p^{t}=1$, where $perclof_{M}^{t}=100$ and $perclof_{I}^{t}=100$, this means that $A^{t}$ is an \textit{extreme outlier} instance and it is safe to create a cloud of artificial samples around it as much as possible. However, as $p^{t}$ decreases, the chance of generating artificial samples around $A^{t}$ declines, because an \textit{extreme outlier} instance attracts more artificial samples around it. For example, if $p^{t}=0.25$, this may map to two cases: either $high \ lof_{M}^{t}$ (close to 1) and $low \ lof_{I}^{t}$ (close to 0), then $A^{t}$ is surrounded by minority instances and we give a lower chance to generating artificial samples around $A^{t}$; or $low \ lof_{M}^{t}$ (close to 0), then $A^{t}$ is surrounded by majority instances and we should seek to generate as few as possible artificial samples around it. In this way, the minority instances having higher peculiarity $p^{t}$ are given more chance to generate artificial samples surrounding them.

\subparagraph{\emph{Artificial Sampling At Feature Level}}
Let $n^{t'}_{f}$ represent the value of the $f^{th}$ feature of a majority instance $N^{t'} \in M$ at the session slot $t'$, and let $a^{t}_{f}$ represent the value of the $f^{th}$ feature of a minority instance $A^{t} \in I_{c}$ at the session slot $t$. For each feature $f; 1 \preceq f \preceq m$, we calculate the distance from $a^{t}_{f}$ of $A^{t} $ to the nearest neighbour $n^{t'}_{f}$ of $N^{t'}$ at the level of feature $f$. In other words, we search the set of majority instances $M$ at the level of feature $f$ only, and we find the closest feature $n^{t'}_{f}$ for $a^{t}_{f}$. At the level of feature $f$, there exists two directions: positive ($+ve$), and negative ($-ve$). Thus, $a^{t}_{f}$ may have (1) only a $+ve$ nearest neighbour, (2) only a $-ve$ nearest neighbour, or (3) both a $+ve$ nearest neighbour and a $-ve$ neatest neighbour. For instance, if $a^{t}_{f}$ is greater than $n^{t'}_{f}$, then the nearest neighbour's distance $distN \text{=} dist(a^{t}_{f},n^{t'}_{f})$ will be positive. On the other hand, if $a^{t}_{f}$ is less than $n^{t'}_{f}$, then $distN$ will be negative. The value(s) of $distN$ for $a^{t}_{f}$ is required to generate \textit{artificial feature values} as demonstrated in Equation \ref{eq:2}.

\begin{figure}[!t]
\scriptsize
\captionsetup{font=scriptsize}
\begin{subfigure}[t]{.47\textwidth}
	\centering
    \includegraphics[width=\textwidth]{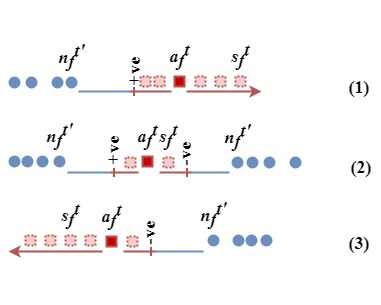}
    \caption{Over one feature (one dimension).}  
    \label{fig:AMOTRE_dim1}
\end{subfigure} 
\hfill
\begin{subfigure}[t]{.52\textwidth}
    \centering
    \includegraphics[width=\textwidth]{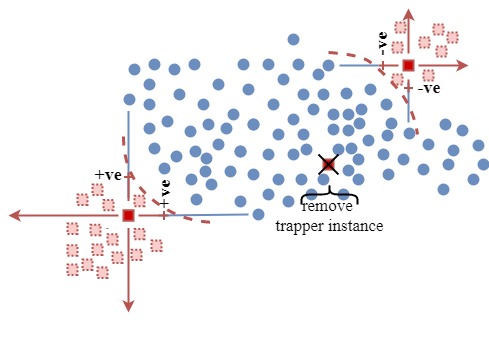}
    \caption{Over two features (two dimensions).} 
    \label{fig:AMOTRE_dim2}
\end{subfigure} 
\caption{Cases of generating artificial insider threat samples in AMOTRE.}  
\label{fig:AMOTRE_dims}
\end{figure}

We define $prob^{+}$ as the probability of generating an artificial sample in the positive ($+ve$) direction. A probability distribution $D_{prob^{+}}$ is devised using the probability $prob^{+}$ to determine whether the direction $dirS$ of the artificial feature value $s^{t}_{f}$ is $+1$ or $-1$. We define $dirS \sim  D_{prob^{+}}$, where $prob^{+}$ is a continuous range of the values of $prob^{+}$. Figure \ref{fig:AMOTRE_dim1} illustrates generating an artificial feature value $s^{t}_{f}$ for $a^{t}_{f}$ at the level of feature $f$. Let a blue circle represent a majority feature value (i.e. feature value $n^{t'}_{f}$ of a majority instance), a red square represent a minority feature value (i.e. feature value $a^{t}_{f}$ of a minority instance), and a red dashed square represent an artificial feature value (i.e. feature value $s^{t}_{f}$ of an artificial instance) generated. In the following, we describe the three cases for generating an artificial feature value:

\begin{itemize}[noitemsep]

\item If $a^{t}_{f}$ has only a $+ve$ nearest neighbour $n^{t'}_{f}$ as Figure \ref{fig:AMOTRE_dim1}(1), then the probability of generating an artificial feature value $s^{t}_{f}$ in the $+ve$ direction is lower. We set up $prob^{+} = 0.2$. The rationale behind this is to give higher chance for generating artificial feature values on the opposite direction of the $+ve$ nearest neighbour (i.e. $-ve$ direction), thus creating a cloud of artificial minority instances away from the nearest majority instances. If the chosen direction $dirS = +1$ based on $D_{prob^{+}}$, then $s^{t}_{f}$ is calculated according to Eq. \ref{eq:2} such that $\lambda = 0.3$. Otherwise, $\lambda = 1$.

\item If $a^{t}_{f}$ has only a $-ve$ nearest neighbour $n^{t'}_{f}$ as Figure \ref{fig:AMOTRE_dim1}(3), then the probability of generating an artificial feature value $s^{t}_{f}$ in the $+ve$ direction is higher. We set up $prob^{+} = 0.8$. The rationale behind this is to give higher chance for generating artificial feature values on the opposite direction of the $-ve$ nearest neighbour (i.e. $+ve$ direction), thus creating a cloud of artificial minority instances away from the nearest majority instances. If the chosen direction $dirS=+1$, then $s^{t}_{f}$ is calculated according to Eq. \ref{eq:2} such that $\lambda = 1$. Otherwise, $\lambda = 0.3$. 

\item If $a^{t}_{f}$ has two nearest neighbours in both directions, as in Figure \ref{fig:AMOTRE_dim1}(2), then the probability of generating an artificial feature value $s^{t}_{f}$ in the $+ve$ and $-ve$ direction is equal. We set up $prob^{+} = 0.5$. This means both directions have equal chance of generating an artificial feature value. If the chosen direction $dirS = +1$ or $dirS = -1$, $s^{t}_{f}$ is calculated according to Eq. \ref{eq:2} such that $\lambda = 0.3$.

\end{itemize}

\begin{equation}
\label{eq:2}
s^{t}_{f} = a^{t}_{f} \ + dirS \times rand(0:\lambda \times distN); s^{t}_{f} \geq 0
\end{equation}
where $\lambda$ represents the parameter that controls the distance permitted to generate artificial feature values along the segment joining $a^{t}_{f}$ and $n^{t'}_{f}$; and $rand(0:\lambda \times distN)$ represents a random number generated to specify the location of the artificial sample along the segment joining $a^{t}_{f}$ and $n^{t'}_{f}$ projected on the feature $f$. $\lambda$ is set up to the value of $0.3$, when the direction chosen to generate an artificial feature value is the same as the direction of the nearest neighbour. The rationale behind this is to locate the artificial feature values on the segment joining $a^{t}_{f}$ and $n^{t'}_{f}$ away from the majority nearest neighbour $n^{t'}_{f}$, thus shielding the majority instances. Otherwise, $\lambda = 1$, such that an artificial feature value can be located along the whole segment. 

Recall that all the artificial feature values in the generated community behaviour data sets are normalised to the range $\left[  0, 1 \right] $. Hence, if the calculated $s^{t}_{f} < 0$, we assign $s^{t}_{f} = min_{f}$ to avoid negative feature values. $min_{f}$ represents the minimum value of the $f^{th}$ feature among the minority instances. 

\subparagraph{\emph{Artificial Sampling At Instance Level}}
Recall that if a minority instance $A^{t}$ is given the chance to be sampled at an iteration (based on its peculiarity), then an artificial sample (instance) $S^{t}$ is generated with a sampling process at the feature level. In other words, for each feature $f; 1 \preceq f \preceq m$, an artificial feature value $s^{t}_{f}$ is generated. Accordingly, an artificial sample (instance) $S^{t} \text{=} \lbrace s^{t}_{1}, s^{t}_{2}, ..., s^{t}_{f} \rbrace$ associated with the minority instance $A^{t}$ is generated. 

As previously mentioned, the steps of Section \ref{sec:3,subsec:2,subsubsec:2} are repeated for a number of iterations until $numS$ of artificial samples are generated. Recall that $I_{r}$ represents the set of remaining minority instances after removing \textit{trapper instances}. Let $I_{s}$ represent the set of artificial samples (instances) generated for $I_{r}$ using the AMOTRE technique. At each iteration, the generated $S^{t}$ is appended to $I_{s}$ ($I_{s} \text{=} I_{s} \cup S^{t}$ as formalised in Algorithm \ref{alg:artificialsampling}). Consequently, the original set of minority instances $I$ is updated to $I\text{=} I_{r} \cup I_{s}$ to comprise the set of remaining minority instances $I_{r}$ (excluding \textit{trapper instances}) and their generated minority samples $I_{s}$.

Figure \ref{fig:AMOTRE_dim2} demonstrates the idea of generating artificial samples over two features (two dimensions). Let a blue circle represent a majority instance $N^{t'}$, a red square represent a minority instance $A^{t}$, and a red dashed square represent an artificial sample (instance) $S^{t}$. It is evident that the farther the minority instance $A^{t}$ from the majority instances as well as other minority instances, the more chance is given to generate artificial samples around it. Furthermore, the artificial samples are mostly generated in the opposite direction of the nearest neighbours from the majority instances, thus shielding the border of the majority instances. 

\vspace{-1em}

\subsection{Class Decomposition Component}
\label{sec:3,subsec:3}
As described previously, the availability of data of both `normal' class and `anomalous class' shapes the insider threat problem as a supervised classification problem. However, the challenge here lies in the data with class imbalance, where the normal instances dominate the minor number of anomalous instances (i.e. malicious insider threats). The performance of a classification method typically tends to decline when the data distributes in an imbalanced way. 

Previous work in addressing class imbalance has considered weakening the effect of the majority class by undersampling its instances \cite{batista2004study,kubat1997addressing,yoon2005unsupervised} -- a process that leads to loss of information, and the possibility of degradation of classification performance as a consequence. Other previous work \cite{jo2004class,japkowicz2001concept} has considered the idea of clustering to guide the sampling process (referred to as cluster-based sampling). The latter tackles the within-class imbalance and further clusters the minority class before oversampling. However, this process can be ineffective in the insider threat problem due to the scarcity and sparsity of the anomalous behaviours.

Vilalta et al. \cite{vilalta2003class} proposed the idea of class decomposition to address the problem of high bias and low variance in the classification methods. The idea of class decomposition tackles the class that distributes in a complex way and applies clustering to this class to decompose it into multiple clusters, thus identifying local patterns within the class. The data of the original class label is assigned the corresponding cluster label as a preprocessing step for the classification method. In this way, the classifier learns multiple decision boundaries (per cluster), rather than a single decision boundary with respect to the original class, and thus avoids data overfitting.

In this work, we adopt class decomposition to address the problem of class imbalance data. Although class decomposition was originally proposed to reduce high bias in classifiers \cite{vilalta2003class}, it has the property of weakening the effect of the decomposed class when constructing the classification model. Such a property is useful if class decomposition is applied to the majority class to address the class imbalance problem. The idea is to decompose the majority class into clusters (i.e. subclasses) to weaken the effect of the majority class with respect to the minority class.

There is a considerable body of literature that has successfully applied $k$-means clustering for class decomposition and/or class imbalance data \cite{banitaan2015class,brodinova2017clustering,japkowicz2001concept,jo2004class,wu2007local}. In the class decomposition component, We select $k$-means clustering \cite{hartigan1979algorithm} to decompose the class into $k$ clusters (subclasses). $k$-means clustering is known to be fast, robust, and less computationally demanding. It only requires tuning the parameter $k$ which controls the number of clusters. Favourably, it does not require, and therefore is not influenced by, data-dependent parameters. In this way, the two-class classification problem (`normal' class label versus `anomalous' class label) is transformed into a multi-class classification problem (labels of clusters (sub-classes) of `normal' class versus `anomalous' class). 

Hence, class decomposition allows the classification method to delineate multiple decision boundaries instead of one decision boundary and to achieve better separation between the majority subclasses and the minority class, thus improving the prediction of new instances.  

\vspace{-1em}

\subsubsection{Decomposing the Majority Class}
\label{sec:3,subsec:3,subsubsec:1}
As described above, we adopt the idea of class decomposition in the insider threat problem to mitigate the bias towards the majority (normal) class in the classification. We apply $k$-means clustering method to decompose the  majority class into $k$ clusters (subclasses). 

Let $X^{t}=\lbrace x^{t}_{1}, x^{t}_{2}, ..., x^{t}_{m} \rbrace$ represent the feature vector at session slot $t$, where $x^{t}_{f}; 1 \preceq f \preceq m$ represents the value of the $f^{th}$ feature. Let $Y=\lbrace y^{1}, y^{2} \rbrace$ represent the output space, where $y^{1}$ is the majority class label and $y^{2}$ is the minority class label. Each instance (i.e. feature vector) $X^{t}$ belongs to a class label $y^{j}, j=\lbrace 1,2 \rbrace$.

Let $M =  X^{t} \  \forall t; X^{t} \in y^{1}$ represent the set of instances that belong to the majority class $y^{1}$, and let $I$ represent the set of instances that belong to the minority class $y^{2}$. If we apply $k$-means clustering method on the set $M$, then the class label $y^{1}$ will break down into $k$ cluster labels. Hence, each instance $X^{t}$ in the set $M$ will be assigned a cluster label instead of a class label. Let $\lbrace yc^{1}_{1}, yc^{1}_{2}, ..., yc^{1}_{k} \rbrace$ represent the set of cluster labels belonging to class label $y^{1}$, where $yc^{1}_{k}; 1 \preceq c \preceq k$ represents the $c^{th}$ cluster label.

\begin{figure}[!t]
\scriptsize
\captionsetup{font=scriptsize}
	\centering
	\includegraphics[width=.4\textwidth]{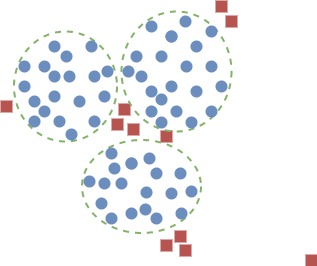}
	\caption{Decomposition of the majority class into $k\text{=}3$ clusters.}
	\label{fig:clustering}
\end{figure}

Figure \ref{fig:clustering} illustrates the idea of applying class decomposition on the majority class. Let the blue circles represent the majority instances (i.e. normal behaviours), and let the red squares represent the minority instances (i.e. anomalous behaviours). The green dashed circles represent the $k\text{=}3$ clusters (patterns) identified among the majority data. In this case, the problem is defined as a multi-class (\textit{four-class}) classification problem, where the output space is represented as $Y=\lbrace  yc^{1}_{1}, yc^{1}_{2}, yc^{1}_{3} , y^{2} \rbrace$, such that $YC = \lbrace yc^{1}_{1}, yc^{1}_{2}, yc^{1}_{3} \rbrace$ denotes the cluster labels, and $y^{2}$ denotes the minority class label.

\vspace{-1em}

\subsubsection{Decomposing the Minority Class}
\label{sec:3,subsec:3,subsubsec:2}
As previously mentioned, we proposed an oversampling technique to tackle the minority class. As later explained in the experiments, the concept of class decomposition may be also applied to the minority class, however, after oversampling its instances. The original size of the minority class is very small, where the number of minority instances (e.g. $132$ anomalous behaviours in com-I) is much lower than the number of majority instances (e.g. $2964$ normal behaviours in com-I) -- \textit{we refer to this as `scarcity'}. Also, the distribution of minority instances is dispersed, where the minority instances may exist among the majority behaviours or may be dispersed among the whole data -- \textit{we refer to this as `sparsity'}. The scarcity and sparsity of the minority instances in the insider threat data sets makes it difficult to cluster the original minority class $I$. If we oversample the set $I$, then the updated set $I$ will append the artificial set $I_{s}$. Eventually, the updated set $I$ can then be decomposed into $k$ clusters.

Consider $I =  X^{t} \  \forall t; X^{t} \in y^{2}$, the set of instances that belong to the minority class $y^{2}$. If we apply $k$-means clustering method on the set $I$, then the class label $y^{2}$ will break down into $k$ cluster labels. Hence, each instance $X^{t}$ in the set $I$ will be assigned a cluster label instead of a class label. Let $\lbrace yc^{2}_{1}, yc^{2}_{2}, ..., yc^{2}_{k} \rbrace$ represent the set of cluster labels belonging to class label $y^{2}$, where $yc^{2}_{k}; 1 \preceq c \preceq k$ represents the $c^{th}$ cluster label.

\vspace{-1em}

\subsection{Classification Component}
\label{sec:3,subsec:4}
Typically, the knowledge discovery system consists of the feature engineering component and the classification component. But, given the challenge of insider threat problem and the problem of class imbalance, we proposed an opportunistic knowledge discovery system AnyThreat with an extension of two components: an oversampling component, and a class decomposition component. 

Typically, the classification component applies a classification method to delineate a decision boundary between the majority class and the minority class -- two-class classification. In the proposed AnyThreat system, the classification method is applied on the subclasses (i.e. clusters) produced by the class decomposition component. Thus, it delineates multiple decision boundaries instead of one boundary to achieve better separation between the majority class(es) (i.e. sublclasses) and the minority class(es) --  multi-class classification. Hence, it improves the prediction of the new instances.

\vspace{-1em}
\section{Experiments}
\label{sec:4} 
In this section, we give a description of the variety of experiments carried out to evaluate the performance of the proposed approach, together with an explanation of the values used for parameter tuning. After that, we introduce the default versions and the refined versions of the evaluation measures utilised to evaluate the performance of the methods. Last but not least, we discuss the results of the experiments with a statistical significance test, and prove the merit of the proposed opportunistic AnyThreat system.

\vspace{-1em}

\subsection{Experimental Setup}
\label{sec:4,subsec:1}
In order to evaluate the effectiveness of the proposed approach, we performed a variety of experiments on the CMU-CERT data sets on Windows Server 2016 on Microsoft Azure (RAM $140GB$, OS $64-bits$, CPU  Intel Xeon $E5-2673 v3$). First, MATLAB $R2016b$ was used to preprocess the data set and generate community data sets per session slots of $4$ hours. We implemented AMOTRE and carried out the experiments in $R$ environment ($R-3.4.1$) using \textbf{Rlof} package for LOF in AMOTRE, \textbf{DMwR} package for SMOTE, \textbf{caret} package \cite{kuhn2008caret} for the classification methods and their evaluation, and \textbf{MASS} package for the Wilcoxon Ranked test.

\vspace{-1em}

\subsection{Description of the Experiments}
\label{sec:4,subsec:2}
Table \ref{tab:exp} presents the variety of experiments carried out to evaluate the effectiveness of the extended components (oversampling component + class decomposition component) on the performance of the proposed opportunistic AnyThreat system. Default represents a base model that only applies one of the classification methods. SMOTE represents a base model that employs SMOTE in the oversampling component and drops the class decomposition component. Similarly, AMOTRE represents a base model that employs the proposed AMOTRE in the oversampling component and drops the class decomposition component. AMO-na represents an experiment that employs AMOTRE without the trapper removal (-TRE) method -- AMO- oversampling. CD(M)-SMOTE and CD(M)-AMOTRE represent experiments that employ SMOTE and AMOTRE respectively in the oversampling component and apply Class Decomposition (CD) on the majority class $M$ only. CD(MI)-SMOTE and CD(MI)-AMOTRE represent experiments that employ SMOTE and AMOTRE respectively in the oversampling component and apply class decomposition on the majority class $M$ and on the minority class $I$. These experiments are evaluated using the measures defined later: \textbf{TP$_{T}$}, \textbf{FP}, and \textbf{F1 measure}. Note that the procedure of trapper removal is not applied with SMOTE, because it was first introduced as an essential part (-TRE part) in the proposed AMO-TRE technique for the oversampling component. In the experiments, we analyse the influence of the -TRE part on our technique by applying the oversampling: (1) without trapper removal (AMO-na), and (2) with trapper removal (AMO-TRE), as revealed in Table \ref{tab:exp}. We then analyse the influence of introducing class decomposition variations to each of SMOTE and AMOTRE.

Each of the experiments is evaluated on five base classification methods utilised in the classification component: Random Forest (rf); Extreme Gradient Boosting (xgb); Support Vector Machines with linear kernel (svmL), polynomial kernel (svmP), and radial basis function kernel (svmR). 

\begin{table}[!t]
\scriptsize
\captionsetup{font=scriptsize}
\centering
\caption {Definition of Experiments.} 
\label{tab:exp}
\begin{tabular}{l l}
\hline\noalign{\smallskip}
 Default & base classification method \\
 SMOTE & SMOTE minority class $I$ \\ 
 CD(M)-SMOTE & SMOTE + CD of majority class $M$ only \\
 CD(MI)-SMOTE & SMOTE + CD of $M$ and $I$ \\
 AMO-na & AMOTRE $I$ without trapper removal \\ 
 AMOTRE & AMOTRE $I$ with trapper removal \\ 
 CD(M)-AMOTRE & AMOTRE + CD of $M$ only\\  
 CD(MI)-AMOTRE & AMOTRE + CD of $M$ and $I$ \\  
 \noalign{\smallskip}\hline\noalign{\smallskip}
\end{tabular}
\end{table} 

The experiments are tuned for different values of parameters as shown in \\Table \ref{tab:param}. Note that an extensive number of experiments was done to select the presented tuning values for the parameters. The values were selected based on the experiments achieving the best performance in terms of the evaluation measures described below.

Regarding class decomposition, we tuned the number of clusters for $k \text{=} \lbrace 2,4,6 \rbrace$ for both (1) the decomposition of the majority class $M$, and (2) the decomposition of the minority class $I$. However, the results for only $k \text{=} 2$ are reported, due to revealing better performance. The proposed approach was able to detect most of the malicious insider threats in the data sets for $k \text{=} 2$. The literature demonstrated the effectiveness of $k$-means clustering for small values of $k$ when applied for class decomposition \cite{banitaan2015class,jo2004class}. 

As explained in Section \ref{sec:3,subsec:2}, a malicious insider threat comprises a set of anomalous behaviours (instances) carried out by a malicious insider. For instance, in community com-P, we have $17$ malicious insider threats. These malicious insider threats are associated to $366$ anomalous instances (behaviours) that make up the minority class $I$. Thus, when applying the class decomposition of the minority class $I$, we are actually clustering the anomalous training instances from the $366$ instances; not from the $17$ malicious insider threats.

Regarding SMOTE technique, the oversampling percentage is tuned for $perc.over$ $\text{=} \lbrace 200,300,400 \rbrace$ to assess whether increasing the percentage of generated artificial samples to the minority class $I$ improves the performance of SMOTE. Similarly, our AMOTRE technique is tuned for $perc.over \text{=} \lbrace 200,300,400 \rbrace$ to generate an equal number of artificial samples to that generated in SMOTE. Note that we report the results for only $perc.over \text{=} 200$, due to achieving the lowest number of false positives (FP); which is the ultimate aim of our work. 

The -TRE part is tuned for only $\tau \text{=} 10$ to test the influence of removing trapper instances from the minority class instances on the overall performance of AMOTRE. We select the survival threshold $\tau \text{=} 10$ so that each minority instance $A^{t} \in I$ having a percentile rank $perclof_{M}^{t} < \tau$ is considered as a trapper instance and removed (as detailed in Section \ref{sec:3}). 

The details about tuning $prob_{+}$ and $\lambda$, for the displayed values in Table \ref{tab:param}, can be found in the description of AMOTRE technique in Section \ref{sec:3}.

\begin{table}[!t]
\scriptsize
\captionsetup{font=scriptsize}
\centering
\caption {Tuned Parameters for SMOTE and AMOTRE.} 
\label{tab:param}
\begin{tabular}{l l}
\hline\noalign{\smallskip}
 $k \text{=} \lbrace 2,4,6 \rbrace$ & number of clusters \\
 $perc.over \text{=} \lbrace 200,300,400 \rbrace$ & percentage of oversampling \\
 $\tau\text{=}10$ & survival threshold for $A^{t}$ in -TRE part \\
 $prob^{+}\text{=}\lbrace 0.2,0.5,0.8 \rbrace$ & controlled by the direction of $n^{t'}_{f}$\\ 
 $\lambda\text{=}\lbrace 0.3,1 \rbrace$ & controlled by $dirS$ and $prob^{+}$\\
 \noalign{\smallskip}\hline\noalign{\smallskip}
\end{tabular}
\end{table}

Based on the challenge of the insider threat problem, a 2-fold cross validation is applied so that the base classification methods learn on a 50\% subsample of the data, and test on a 50\% subsample repetitively for $2$ times. This allows the 50\% subsample testing data to have more instances, including \textit{more anomalous instances} (behaviours). Hence, it helps us to reveal the performance of the proposed opportunistic AnyThreat system. In 2-fold cross validation, each fold may contain a subset of the anomalous behaviours belonging to a malicious insider threat. In terms of the training, this provides `weak supervision', as some of the anomalous behaviours associated to a particular threat will be missing in the training fold. In terms of the testing, this would show the `robustness' of the approach being able to detect the malicious insider threat whose behaviours are partially represented in the training fold, even with a weak signal (threats are partially represented in the test fold). 

\vspace{-1em}

\subsection{Evaluation Measures}
\label{sec:4,subsec:3} 
Much research has been done to detect or mitigate malicious insider threats, but standard measures have not been established to evaluate the proposed models \cite{greitzer2013methods}. The research practices show that the insider threat problem demands the measurement of the effectiveness of the models before being deployed, preferably in terms of true positives (TP) and false positives (FP) \cite{guido2013insider}.  

The variety of the utilised evaluation measures in the state-of-the-art reveals the critical need to formulate the insider threat problem and to define the measures that would best validate the effectiveness of the proposed approach. In the following, we define the evaluation measures utilised in this article.

As previously mentioned, the ultimate aim of this approach is to detect any-behaviour-all-threats (\textit{opportunistic approach} as described in Section \ref{sec:1}), and to reduce the number of false alarms.
In the following, we define the measures used to evaluate the performance. We introduced refined versions of some measures. The rationale behind this is related to our ultimate aim, which is to detect the malicious insider threats (not necessarily all anomalous behaviours per threat), and at the same time to reduce FPs (all false alarms of false predicted behaviours).

\begin{itemize}[noitemsep]
\item P: \textit{Positives} number of anomalous instances (anomalous behaviours);
\item P$_{T}$: \textit{Threats} number of malicious insider threats associated to anomalous instances. In other words, P$_{T}$ is the number of malicious insiders attributed to the anomalous behaviours;
\item \textbf{TP$_{T}$}: \textit{True Positives} a refined version of the default TP to evaluate the number of threats detected by the system among all the P$_{T}$ malicious insider threats. TP$_{T}$ is incremented if at least one anomalous instance (behaviour associated to the threat) is predicted as anomalous;
\item \textbf{FP}: \textit{False Positives} number of normal instances (behaviours) that are detected as anomalous instances;
\item TN: \textit{True Negatives} number of normal instances (behaviours) that are predicted as normal;
\item FN$_{T}$: \textit{False Negatives} a refined version of the default FN to evaluate the number of insider threats not detected; and
\item \textbf{F1} measure: defined based on the values of the above defined measures. Note that F1 is not close to $1$ due to the use of refined versions of some measures, but this does not reflect low performance.
That is because the maximum TP$_{T}$ (evaluated per threat) is much lower than the minimum FP (evaluated per behaviour). 
\end{itemize}

\vspace{-1em}

\subsection{Results and Discussion}
\label{sec:4,subsec:4}
In this Section, we discuss the results of the variety of experiments carried out, and show the merit of the proposed AnyThreat system according to the following objectives:

\begin{enumerate}[noitemsep]
\item assessing the performance of the AnyThreat system in terms of TP$_{T}$ and FP measures;
\item assessing the influence of trapper removal on the proposed AMOTRE in the oversampling component in terms of F1 measure; and
\item assessing the effectiveness of introducing the class decomposition component for the majority class and the minority class.
\end{enumerate}

\subsubsection{Assessment of AnyThreat in terms of TP$_{T}$ and FP Measures.}
\label{sec:4,subsec:4,subsubsec:1}

The following will address the TP$_{T}$ and FP measures, which represent key measures for the insider threat problem. 

Table \ref{tab:maxTP} reports the maximum TP$_{T}$ in each of the experiments over the communities, associated with the base classifier(s) which achieved the maximum TP$_{T}$. 

Over com-P, Table \ref{tab:maxTP} shows that the Default detects only TP$_{T}\text{=}11/16$, missing $5$ malicious insider threats. Note that $11/16$ represents the number of detected threats TP$_{T}\text{=}11$ out of the number of threats P$_{T}\text{=}16$. On the other hand, CD(M)-SMOTE and CD(M)-AMOTRE achieve the maximum TP$_{T}\text{=}14/16$, where only $2$ malicious insider threats are not detected.

Over com-S, the Default detects TP$_{T}\text{=}17/21$, missing $4$ malicious insider threats. CD(MI)-SMOTE, AMOTRE and CD(MI)-AMOTRE attain the maximum TP$_{T}$, where CD(MI)-SMOTE detects TP$_{T}\text{=}17/20$; while AMOTRE and CD(MI)-AMOTRE detect TP$_{T}\text{=}18/21$. Only $3$ malicious insider threats are missed.

Over com-I, the Default detects TP$_{T}\text{=}7/11$, missing $4$ malicious insider threats. SMOTE attains the maximum TP$_{T}\text{=}9/12$, where only $3$ malicious insider threats are missed.

It is evident that the Default failed to detect as much malicious insider threat as the other experiments of the proposed AnyThreat system over all the communities. The maximum TP$_{T}$ is achieved by the variations of class decomposition experiments with AMOTRE and SMOTE over the communities, excluding com-I where SMOTE of the oversampling component achieved the maximum TP$_{T}$. Recall that the ultimate aim in the insider threat problem is to detect \textbf{all} malicious insider threats. Hence, the our AnyThreat system with the proposed components demonstrate better performance in terms of TP$_{T}$ measure.

\begin{table}[!t]
\scriptsize
\captionsetup{font=scriptsize}
\centering
\caption {Maximum TP$_{T}$/P$_{T}$ of detected insider threats over communities associated with the base classifier(s) which achieved the maximum TP$_{T}$.}
\label{tab:maxTP}
\begin{tabular}{ l | l | l | l}
\hline\noalign{\smallskip}
 Experiment & com-P & com-S & com-I \\ 
\hline\noalign{\smallskip}
 Default & 11/16 $\lbrace$rf,xgb$\rbrace$ & 17/21 $\lbrace$rf,xgb$\rbrace$ & 7/11 rf \\
 SMOTE & 13/16 $\lbrace$xgb,svmL$\rbrace$ & 17/21 svmR & \textbf{9/12} xgb\\ 
 CD(M)-SMOTE & \textbf{14/16}  svmR & 17/22 $\lbrace$xgb,svmL$\rbrace$ & 8/12 $\lbrace$rf,xgb$\rbrace$\\
 CD(MI)-SMOTE & 13/16 svmL & \textbf{17/20} $\lbrace$xgb,svmP$\rbrace$ & 8/12 rf\\
 AMOTRE & 13/16 $\lbrace$svmL,svmP$\rbrace$ & \textbf{18/21} svmL & 8/12 $\lbrace$svmP,svmR$\rbrace$\\ 
 CD(M)-AMOTRE & \textbf{14/16} $\lbrace$svmL,svmR$\rbrace$ & 17/22 $\forall \setminus$ svmR & 8/12 rf\\ 
 CD(MI)-AMOTRE & 13/16 $\lbrace$svmL,svmP$\rbrace$ & \textbf{18/21} $\lbrace$rf,xgb,svmL$\rbrace$ & 8/12 $\lbrace$rf,svmL$\rbrace$\\
 \noalign{\smallskip}\hline\noalign{\smallskip}
\end{tabular}
\caption*{$\forall$ $\vert$ for all base classifiers\\ $\forall\setminus$xxxx $\vert$ for all base classifiers except xxxx\\ svm- $\vert$ for all utilised SVM methods}
\end{table}

\begin{table}[!t]
\scriptsize
\captionsetup{font=scriptsize}
\centering
\caption {Minimum FP over communities associated with the base classifier(s) which achieved the minimum FP.}
\label{tab:minPercFP}
\begin{tabular}{ l | l | l | l }
\hline\noalign{\smallskip}
 Experiment & com-P & com-S & com-I\\ 
\hline\noalign{\smallskip}
 base classifier & 0 $\lbrace$svmL,svmR$\rbrace$ & 20 svmR & 0 $\lbrace$svmL,svmR$\rbrace$ \\
 SMOTE & 69 xgb & 117 xgb & 5 svmR\\ 
 CD(M)-SMOTE & 80 xgb & 123 rf & 4 svmR\\
 CD(MI)-SMOTE & 80 rf & 118 rf & \textbf{0} svmP\\
 AMOTRE & 49 xgb & \textbf{88} xgb & 3 svmR\\ 
 CD(M)-AMOTRE & 47 rf,xgb & 101 xgb & 3 svmR\\  
 CD(MI)-AMOTRE & \textbf{46} rf & \textbf{88} xgb & 6 svmR\\
 \noalign{\smallskip}\hline\noalign{\smallskip}
\end{tabular}
\end{table}

To test the significance of the results, we use the Wilcoxon Signed-Rank Test which compares each pair of experiments. The Default base classifiers are compared to the proposed AnyThreat system in terms of the TP$_{T}$ measure, as the ultimate aim of the opportunistic approach is to detect all malicious insider threats. The p-value for $\langle$Default, AnyThreat$\rangle$ is $0.001069 < .05$. Hence, the proposed AnyThreat system is significantly different from Default base classifiers at $.05$ significance level. 

In Table \ref{tab:minPercFP}, we report the minimum FP measure in each experiment over the communities associated with the base classifier(s) which achieved the minimum FP.

Over com-P, Table \ref{tab:minPercFP} CD(MI)-AMOTRE reduces the number of FPs to the minimum FP$\text{=}46$ ($3.36$\%). Over com-S, AMOTRE and CD(MI)-AMOTRE attain the minimum FP$\text{=}88$. Over com-I, CD(MI)-SMOTE attains the minimum FP$\text{=}0$; svmP, compared to a minimum FP$\text{=}3$ for AMOTRE and CD(M)-AMOTRE.
 
On the other hand, the results show that the Default attains FP$\text{=}0$ over com-P and com-I, however, the Default which attained FP$\text{=}0$ actually did not detect any malicious insider threat (TP$_{T}\text{=}0$). Similarly, the Default which attained FP$\text{=}20$ over com-S detected only TP$_{T}\text{=}6/21$ malicious insider threat; missing $15$ malicious insider threats.

Given our ultimate aim to detect all malicious insider threats, the Default base classifier(s) fail to prove the best performance in terms of the FP measure. Hence, the proposed variations of class decomposition experiments with AMOTRE and SMOTE achieve the minimum number of FP over all the communities, while detecting most of the malicious insider threats.

In conclusion, the integration of proposed oversampling component (SMOTE or AMOTRE) and the class decomposition component (the variations of class decomposition) demonstrate the best performance in terms of TP$_{T}$ and FP measures compared to the Default base classifiers (objective (1)). Hence, this emphasises the importance of integrating the proposed components in the proposed AnyThreat system.

\vspace{-1em}

\subsubsection{Influence of Trapper Removal on the Proposed AMOTRE in the Oversampling Component} 
\label{sec:4,subsec:4,subsubsec:2}
The proposed technique AMOTRE consists of two key parts: the oversampling part (AMO-), and the trapper removal part (-TRE). In the following, we assess the influence of the -TRE part on the AMOTRE technique. Table \ref{tab:compareexperiments} provides the values of F1 measure for AMO-na (AMO- without trapper removal) and AMOTRE (AMO-TRE with trapper removal) on the minority class $I$ . 

Over com-P, AMOTRE reports a higher F1 measure than AMO-na for all classifiers except svmP and svmR. AMOTRE attains the maximum F1$\text{=}0.2894$;xgb compared to a maximum F1$\text{=}0.2637$;xgb for AMO-na. Over com-S, AMOTRE also reports a higher F1 than AMO-na for all classifiers except xgb. AMOTRE attains a maximum F1$\text{=}0.2517$;xgb compared to the maximum F1$\text{=}0.2556$;xgb for AMO-na. Over com-I, AMOTRE reports a higher F1 than AMO-na for all classifiers. AMOTRE attains the maximum F1$\text{=}0.6956$;svmR compared to a maximum F1$\text{=}0.5714$;svmR for AMO-na. Note that AMO-na shows equal performance to AMOTRE in terms of F1 with respect to $\lbrace$svmL,svmP$\rbrace$.

We can conclude that removing \textit{trapper instances} from the minority class $I$ boosts the performance of the base classifier in terms of F1 measure (objective (2)). These trapper instances, if not removed, would be selected in the AMOTRE iterations to generate artificial samples around them. This would trap the classifier from finding the optimal decision boundary that separates majority instances from minority instances, which in turn results in a high number of false alarms. However, removing trapper instances is a precautionary step towards reducing the number of FPs, and achieving a higher F1 measure.

\vspace{-1em}

\subsubsection{Effectiveness of Introducing the Class Decomposition Component}
\label{sec:4,subsec:4,subsubsec:3}
Here, we assess the effectiveness of class decomposition component on the proposed AnyThreat system. Table \ref{tab:compareexperiments} provides the values of F1 measure for the variations of class decomposition experiments compared to the base models of SMOTE and AMOTRE over the communities in 2-fold cross validation. The results are reported with respect to the five base classifiers.

\begin{table}[!t]
\scriptsize
\captionsetup{font=scriptsize}
\centering
\caption {Comparing the values of F1 measure for each of the defined experiments over the communities in 2-fold cross validation. The results are reported with respect to the five base classifiers $\lbrace$rf, xgb, svmL, svmP, svmR$\rbrace$. The values of F1 measure is associated with the TP$_{T}$ achieved for the Default base classifiers.}
\label{tab:compareexperiments}

\begin{subtable}{.98\textwidth}
\centering
\begin{tabular}{ l | l | l | l | l | l }
\hline\noalign{\smallskip}
 Experiment & rf & xgb & svmL & svmP & svmR \\
\hline\noalign{\smallskip}
 base classifier & 0.3013 (11/16) & 0.2558 (11/16) & / (0/16) & 0.3703 (5/16) & / (0/16)\\
 SMOTE & 0.2142 & 0.2653 & 0.1452 & 0.1256 & 0.1454 \\
 AMOTRE & 0.2558 & 0.2894 & 0.1699 & 0.1969 & 0.1860 \\ 
 AMO-na & 0.2222 & 0.2637 & 0.1677 & 0.2295 & 0.1925 \\
 CD(M)-SMOTE & 0.2222 & 0.2056 & 0.1313 & 0.1165 & 0.1590 \\
 CD(M)-AMOTRE & 0.32 & 0.32 & 0.2089 & 0.2166 & 0.2121 \\
 CD(MI)-SMOTE & 0.2056 & 0.2037 & 0.1405 & 0.1297 & 0.1549 \\
 CD(MI)-AMOTRE & 0.2777 & 0.2597 & 0.2407 & 0.2280 & 0.1739 \\
 \noalign{\smallskip}\hline\noalign{\smallskip}
\end{tabular}
\caption {com-P.}
\label{tab:comPexperiments}
\end{subtable}

\begin{subtable}{.98\textwidth}
\centering
\begin{tabular}{ l | l | l | l | l | l }
\hline\noalign{\smallskip}
 Experiment & rf & xgb & svmL & svmP & svmR \\
\hline\noalign{\smallskip}
 base classifier & 0.2677 (17/21) & 0.2615 (17/21) & 0.3013 (11/21) & 0.2647 (9/21) & 0.2553 (6/21)\\
 SMOTE & 0.1886 & 0.2077 & 0.2073 & 0.1797 & 0.2 \\
 AMOTRE & 0.2463 & 0.24 & 0.2517 & 0.2344 & 0.2191 \\
 AMO-na & 0.2207 & 0.2556 & 0.2312 & 0.2193 & 0.2111 \\
 CD(M)-SMOTE & 0.1987 & 0.2085 & 0.1827 & 0.1711 & 0.1860 \\
 CD(M)-AMOTRE & 0.2236 & 0.2428 & 0.2207 & 0.2312 & 0.2176 \\
 CD(MI)-SMOTE & 0.2179 & 0.2142 & 0.2 & 0.1910 & 0.2073 \\
 CD(MI)-AMOTRE & 0.2834 & 0.2686 & 0.2368 & 0.2377 & 0.2411 \\
 \noalign{\smallskip}\hline\noalign{\smallskip}
\end{tabular}
\caption {com-S.}
\label{tab:comSexperiments}
\end{subtable}

\begin{subtable}{.98\textwidth}
\centering
\begin{tabular}{ l | l | l | l | l | l }
\hline\noalign{\smallskip}
 Experiment & rf & xgb & svmL & svmP & svmR \\
\hline\noalign{\smallskip}
 base classifier & 0.6363 (7/11) & 0.2666 (4/11) & / (0/11) & 0.2857 (2/11) & / (0/11)\\
 SMOTE & 0.4444 & 0.4 & 0.5 & 0.5925 & 0.5833 \\
 AMOTRE & 0.6086 & 0.4285 & 0.4516 & 0.5 & 0.6956 \\
 AMO-na & 0.5384 & 0.3809 & 0.4516 & 0.5 & 0.5714 \\
 CD(M)-SMOTE & 0.3902 & 0.3636 & 0.4615 & 0.4615 & 0.5454 \\
 CD(M)-AMOTRE & 0.5714 & 0.3636 & 0.4516 & 0.461 & 0.6363 \\ 
 CD(MI)-SMOTE & 0.4848 & 0.3333 & 0.5555 & 0.625 & 0.6315 \\
 CD(MI)-AMOTRE & 0.5714 & 0.4666 & 0.5 & 0.48 & 0.56 \\
 \noalign{\smallskip}\hline\noalign{\smallskip}
\end{tabular}
\caption {com-I.}
\label{tab:comIexperiments}
\end{subtable}

\end{table}

Class decomposition is applied in two different strategies: to the majority class $M$ only as in CD(M)-SMOTE and CD(M)-AMOTRE; and to the majority class $M$ as well as to the minority class $I$ as in CD(MI)-SMOTE and CD(MI)-AMOTRE.

Over com-P, CD(M)-AMOTRE outperforms AMOTRE base model in terms of F1 with respect to all base classifiers.
CD(M)-SMOTE outperforms SMOTE base model with respect to r$\lbrace$rf,svmR$\rbrace$. Overall, CD(M)-AMOTRE attains the maximum F1$\text{=}0.32$;$\lbrace$rf,xgb$\rbrace$ compare to all other experiments. With respect to the Default, CD(M)-AMOTRE outperforms all base classifiers except svmP which achieved better F1 measure, however, TP$_{T}\text{=}5/16$;svmP only in this case.

Over com-S, CD(MI)-AMOTRE outperforms AMOTRE base model in terms of F1 with respect to all base classifiers. CD(MI)-SMOTE outperforms SMOTE base model with respect to all base classifiers except svmL. Overall, CD(MI)-AMOTRE attains the maximum F1$\text{=}0.2834$;rf compared to all other experiments. With respect to the Default, CD(MI)-AMOTRE outperforms rf and xgb base classifiers, while Default achieves better F1 measure in with respect svm-. Similarly, in the case Default achieves better F1 measure, the maximum of TP$_{T}\text{=}11/21$;svmL only was achieved; missing $10$ malicious insider threats.

Over com-I, CD(MI)-AMOTRE outperforms AMOTRE base mode with respect to $\lbrace$xgb,svmL$\rbrace$. CD(MI)-SMOTE outperforms SMOTE base model with respect to all base classifiers except xgb. Overall, AMOTRE attains the maximum F1$\text{=}0.6956$;svmR followed with a maximum F1$\text{=}0.6363$;svmR for CD(M)-AMOTRE compared to all other experiments. With respect to the Default, CD(MI)-AMOTRE outperforms all base classifiers except rf which achieved better F1 measure, however, TP$_{T}\text{=}7/11$;rf only in this case.

It is evident that the Default base classifiers fail to prove the best performance in terms of the F1 measure given that it did failed to detect as much malicious insider threats as the other experiments of the proposed AnyThreat system over all the communities (ultimate aim of detecting all threats). Hence, introducing class decomposition along to the oversampling technique (as in CD(M)-SMOTE or CD(MI)-SMOTE and CD(M)-AMOTRE or CD(MI)-AMOTRE) improved the performance of AMOTRE base model and SMOTE base model in terms of F1 measure over all base classifiers, as well as it improved the performance of the Default base classifier(s) in terms of F1 measure and particularly TP$_{T}$.

Accordingly, we can deduce the effectiveness of introducing the class decomposition component to the AnyThreat system (objective (3)).

\vspace{-1em}

\section{Conclusion and Future Work} 
\label{sec:5}
In this article, we address the insider threat problem as a knowledge discovery problem with class imbalance with the aim of detecting all malicious insider threats while reducing the number of false alarms. We propose an opportunistic knowledge discovery system, namely AnyThreat, with the aim to detect any-behaviour-all-threats; we can \textbf{hunt} a malicious insider threat by at least detecting one anomalous behaviour associated to this threat. This will contribute in reducing the false alarms. The AnyThreat system consists of four components: a feature engineering component, an oversampling component, a class decomposition component, and a classification component. We define AMOTRE as a selective oversampling technique for the oversampling component, where it selects the minority instances to be oversampled based on a measured local outlier factor. The minority instances located in a high-density region of majority instances are trapper instances that would trap the classification method. In AMOTRE, trapper instances are removed to reduce the number of false alarms. The role of the class decomposition component is to weaken the majority class by decomposing it into subclasses, so the decision of the classifier would not bias toward the majority class. Class decomposition can also be applied to the minority class after oversampling in order to achieve balance between the subclasses.

We evaluate different variations of applying the state-of-the-art sampling technique SMOTE or the proposed selective oversampling technique AMOTRE in the oversampling component with and without integrating the class decomposition component with respect to five high performing base classification methods. The experiments showed that the \textit{trapper removal} is a key part of AMO-TRE oversampling technique, where AMOTRE reveals higher F1 measure compared to AMO-na with respect to the base classifiers over all communities. Moreover, introducing the variations of class decomposition improved the performance of the base classifiers, thus emphasising the effectiveness of the class decomposition component. The results show the merit of the proposed opportunistic knowledge discovery system AnyThreat in terms of achieving the maximum number of detected threats TP$_{T}$, while reducing the number of FPs compared to the SMOTE abd AMOTRE base models, as well as to the Default base classifiers.

Several directions on utilising alternative methods in the proposed AnyThreat system are open. Future work includes optimising the hyperparameters of the classification methods through metaheuristics methods, such as Genetic Algorithm (GA) for example, which can lead to a higher precision and recall. Also, one can test the proposed AnyThreat system with an alternative renowned oversampling technique other than SMOTE and AMOTRE for the oversampling component. Furthermore, one can adopt the proposed concept of \textit{trapper removal} (-TRE part) in conjunction with a renowned oversampling technique. This would show some empirical evidence of successful hybridisation (e.g. SMOTE-TRE, which is a hybridisation of the -TRE part and the state-of-the-art SMOTE). For the class decomposition component, we can utilise an alternative clustering method for $k$-means clustering. For instance, Nickerson et al. \cite{nickerson2001using} utilised Principal Direction Divisive Partitioning (PDDP) to guide the resampling in imbalanced data sets. PDDP \cite{boley1998principal} determines the internal structure of a class and has a linear complexity of running time.

\bibliographystyle{spmpsci}      
\bibliography{bibliography}   

\end{document}